\documentclass[10pt,twocolumn,letterpaper]{article}
\usepackage{iccv}
\usepackage{times}
\usepackage{epsfig}
\usepackage{graphicx}
\graphicspath{ {images/} }
\usepackage{amsmath}
\usepackage{amssymb}
\usepackage{fancyhdr}
\usepackage[breaklinks=true,bookmarks=false]{hyperref}

\iccvfinalcopy 

\def\iccvPaperID{510} 

\ificcvfinal\fi
\begin{document}

\title{Attentive Semantic Video Generation using Captions}

\author{
Tanya Marwah\thanks{Equal Contribution}\\
IIT Hyderabad\\
{\tt\small ee13b1044@iith.ac.in}
\and
Gaurav Mittal\footnotemark[1]\\
{\tt\small gaurav.mittal.191013@gmail.com}
\and 
Vineeth N. Balasubramanian\\
IIT Hyderabad\\
{\tt\small vineethnb@iith.ac.in}
}


\maketitle
\thispagestyle{empty}


\begin{abstract}
This paper proposes a network architecture to perform variable length semantic video generation using captions. We adopt a new perspective towards video generation where we allow the captions to be combined with the long-term and short-term dependencies between video frames and thus generate a video in an incremental manner. Our experiments demonstrate our network architecture's ability to distinguish between objects, actions and interactions in a video and combine them to generate videos for unseen captions. The network also exhibits the capability to perform spatio-temporal style transfer when asked to generate videos for a sequence of captions. We also show that the network's ability to learn a latent representation allows it generate videos in an unsupervised manner and perform other tasks such as action recognition.\footnote{Accepted at International Conference on Computer Vision (ICCV) 2017 to be held at Venice, Italy.}
\end{abstract}
\vspace{-6pt}
\section{Introduction}
\vspace{-6pt}
What does the mention of a video evoke in a listener's mind? A long sequence of images~(frames) with multiple changing scenes which are temporally and spatially linked to each other. Making a machine generate such an entity is a highly involved task as it requires the machine to learn how to coordinate between the long-term and short-term correlations existing between the various elements of a video. This is the primary motivation behind our work where we take the first steps towards semantic video generation with captions generating variable length videos one frame at a time.

There have been efforts in the recent past which attempt to perform unsupervised video generation~\cite{vondrick2016generating,saito2016temporal} in general, without any specific conditioning on captions. However, from an application perspective, it may not be very useful as there doesn't exist any semantic control over what will be generated at run time. In this work, we propose an approach that attempts to provide this control and streamlines video generation by using captions.

Since a video can be arbitrarily long, generation of such videos necessitates a step-by-step generation mechanism. Therefore, our model approaches video generation iteratively by creating one frame at a time, and conditioning the generation of the subsequent frames by the frames generated so far. Also, every frame is itself an amalgamation of several objects moving and interacting with each other. In order to generate such frames, we follow a recurrent attentive approach similar to~\cite{gregor2015draw}, which focuses on one part of the frame at each time step for generation, and completes the frame generation over multiple time steps. By iteratively generating the frame over a number of time-steps in response to a given caption, our network adds caption-driven semantics to the generated video. A key advantage of following such an approach is the possibility to generate videos with multiple captions and thus change the contents of the video midway according to the new caption.

In order to achieve the aforementioned objectives, the proposed network is required to not only account for the local correlations between any two consecutively generated frames but also has to ensure that the long-term spatio-temporal nature of the video is preserved so that the generated video is not just a loosely coupled set of frames but exhibits a holistic portrayal of the given caption. In this work, we adopt a fresh perspective in this direction, by devising a network that learns these long-term and short-term video semantics separately but simultaneously. To create a frame based on a caption, instead of trivially conditioning the generation on the caption text, we introduce a soft-attention over the captions separately for the long-term and short-term contexts. This attention mechanism serves the crucial role of allowing the network to selectively combine the various parts of the caption with these two contexts and hence, significantly improves the quality of the generated frame. Experiments show that the network, when given multiple captions, is even able to transition between scenes transferring the spatio-temporal semantics it has learned thus far, dynamically during generation.

This paper makes the following contributions: (1) A novel methodology that can perform variable length semantic video generation with captions by separately and simultaneously learning the long-term and short-term context of the video; (2) A methodology for selectively combining information for conditioning at various levels of the architecture using appropriate attention mechanisms; and (3) A network architecture which learns a robust latent representation of videos and is able to perform competently on tasks such as unsupervised video generation and action recognition. The results obtained on standard datasets that earlier efforts have used for such efforts are very promising, and support the claims of this work.

\vspace{-3pt}
\section{Related Work}
\vspace{-6pt}
Recent years have witnessed the emergence of generative models like Variational Autoencoders (VAEs)~\cite{kingma2013auto}, Generative Adversarial Networks (GANs) ~\cite{goodfellow2014generative} and other  methods such as using autoregression~\cite{van2016wavenet}. Earlier generative models based on methods such as Boltzmann Machines (RBMs)~\cite{smolensky1986information,salakhutdinov2009deep} and Deep Belief Nets~\cite{hinton2006fast} were always constrained by issues such as intractability of sampling. Recent methods have enabled learning a much more robust latent representation of the given data~\cite{radford2015unsupervised,radford2015unsupervisedlap,gregor2015draw,zhang2016stackgan} and are helping improving the performance of several supervised learning tasks~\cite{ledig2016photo,yeh2016semantic,xu2015show}.

Previous approaches for image generation have extended VAEs and GANs to generate images based on captions by conditioning them with textual information~\cite{mansimov2015generating,reed2016generative}. Our approach also constitutes a variant of Conditional VAE~\cite{sohn2015learning} but differs in that in our paper, we use it to generate videos (earlier efforts only generated images) from captions with the capability to generate videos of arbitrary lengths. Further, our approach is distinct in that it incorporates captions by learning an attention-based embedding over them by leveraging long-term and short-term dependencies within the video to generate coherent videos based on captions. 

The proposed methodology draws some similarity with past methods~\cite{vondrick2016generating,saito2016temporal} that perform unsupervised video generation (but without captions), with all of them using GANs. Vondrick et al. \cite{vondrick2016generating} use a convolutional network with a fractional stride as the generator and a spatio-temporal convolutional network as a discriminator in a two-stream approach where the background and the foreground of a scene are processed separately. Saito et al. \cite{saito2016temporal} introduced a network called Temporal GAN where they use a 1D deconvolutional network to output a set of latent variables, each corresponding to a separate frame and an image generator which translates them to a 2D frame of the video. However, in addition to not incorporating caption-based video generation, these approaches suffer from the drawback of not being scalable in generating arbitrarily long videos. Our approach, by approaching video generation frame-by-frame and utilizing the long-term and short-term context separately, effectively counters these limitations of earlier work. Besides, our approach learns to focus on separate objects/glimpses in a frame unlike~\cite{kalchbrenner2016video} where the focus is on individual pixels, and ~\cite{vondrick2016generating} where network's attention is divided into just background and foreground.

In addition to generative modeling of videos, learning their underlying representations has been used to assist several supervised learning tasks such as future prediction and action recognition. \cite{srivastava2015unsupervised} is one of the earliest efforts in learning deep representations for videos and utilizes Long Short-Term Memory units (LSTMs)~\cite{hochreiter1997long} to predict future frames given a set of input frames. 
More recent efforts on video prediction have attempted to factorize the joint likelihood of the video to predict the future frames~\cite{kalchbrenner2016video} or use the first image of the video with Conditional VAE to predict the motion trajectories and use them to generate subsequent frames\cite{walker2016uncertain}. We later demonstrate that in addition to the primary application of semantically generating videos with attentive captioning, it is possible to make small changes to our network which enable it to even predict future frames given an input video sequence.

\vspace{-3pt}
\section{Methodology}
\vspace{-6pt}
The primary focus of our work is to allow video generation to take place semantically via captions. 
Generating a video with variable number of frames given a caption is very different from generating a single image for a caption, due to the need to model the temporal context in terms of the actions and interactions among the objects. Consider, for example, the caption -- ``A man is walking on the grass.'' Generating a single image for this caption, ideally, will simply show a man in a walking stance on top of a grassy ground. However, if we want to generate a sequence of frames corresponding to this caption, it necessitates the network to understand how the structure of man on the grass will transition from one frame to the other in relation to the motion of walking. In other words, it means that the network should be able to decouple the fundamental building blocks of any video, i.e., objects, actions and interactions, and have the ability to combine them semantically given a stimulus, which in this case, is a caption. We now explain how we model our approach to generate variable length videos conditioned on captions and later discuss the network architecture.


\begin{figure}[h]
\centering

\includegraphics[width=0.65\columnwidth]{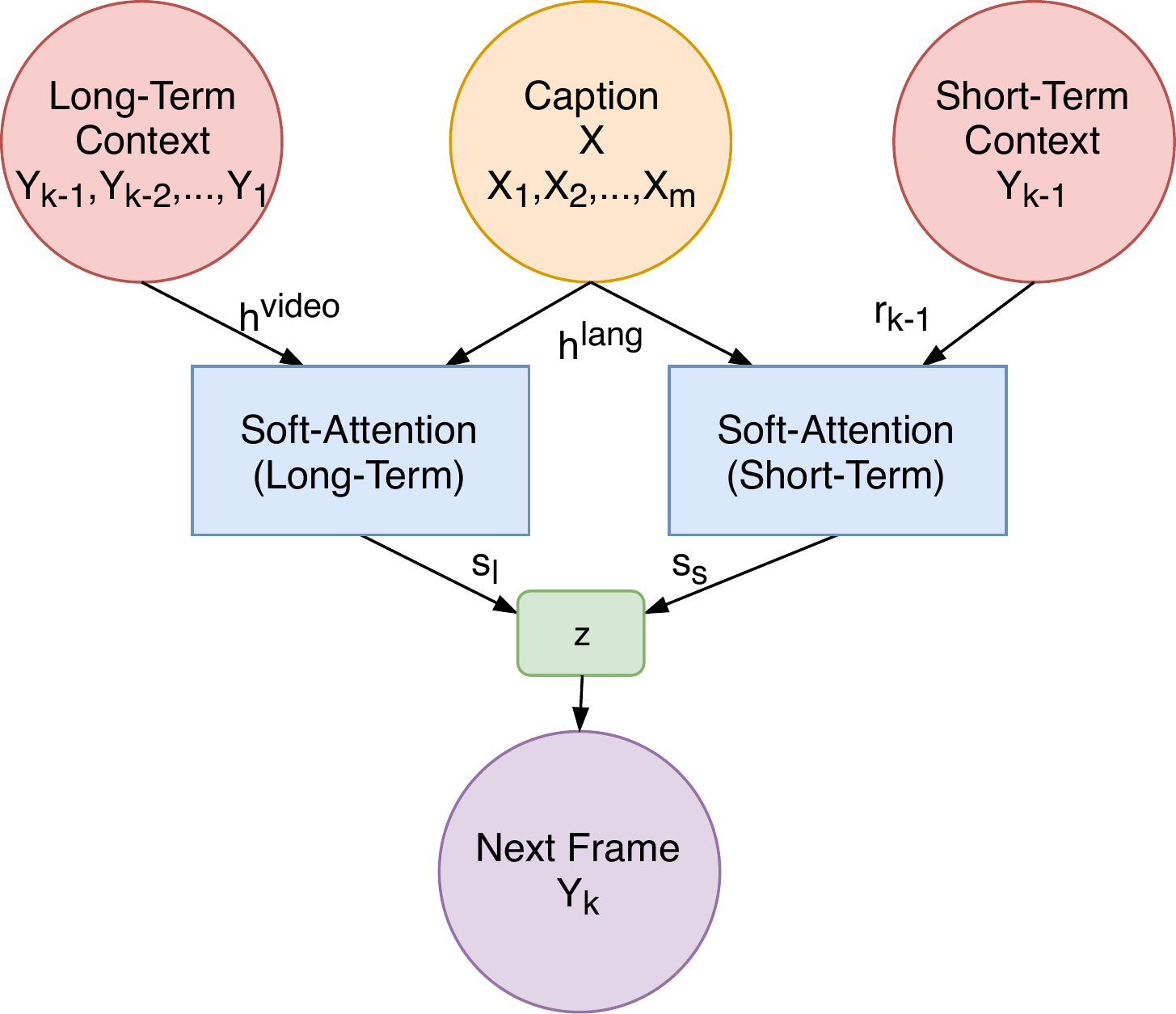}
\caption{Illustration of the proposed model.}
\label{state-space}
\end{figure}
\subsection{Model}
\label{Model}
\vspace{-6pt}
Let the random variable $Y=\{Y_1, Y_2,\cdots, Y_n\}$ denote the distribution over videos with $Y_1, Y_2,\cdots, Y_n$ being the ordered set of $n$ frames comprising the videos. Let $X=\{X_1, X_2,\cdots, X_m\}$ be the random variable for the distribution over text captions with $X_1, X_2,\cdots, X_m$ being the ordered set of $m$ words constituting the caption. $P(Y|X)$ then captures the conditional distribution of generating some video in $Y$ for a given caption in $X$. Our objective is to maximize the likelihood of generating an appropriate video for a given caption. Since we would like to generate a video for a caption one frame at a time, we redefine $P(Y|X)$ as:
\vspace{-6pt}
\begin{equation}
P(Y|X) = \prod^{n}_{i=1} P(Y_i|Y_{i-1},\cdots\cdots,Y_1,X)
\end{equation}
where $n$ is the total number of frames in the video. Generation of the $k^{th}$ frame, $Y_k$, can therefore be expressed as:
\begin{equation}
P(Y_k|X) = P(Y_k|Y_{k-1},\cdots, Y_1, X)
\label{prob-frame}
\end{equation}
thus allowing the generation of a given frame to depend on all the previously generated frames. The generation can hence model both short-term and long-term spatio-temporal context. $Y_k$ gathers short-term context, which consists of the local spatial and temporal correlations existing between any two consecutive frames, from $Y_{k-1}$. $Y_k$ also obtains its long-term context from all the previous frames combined to understand the overall flow of the video. This ensures that that the overall consistency of the video is maintained while generating the frame. In order to model this when generating the $k$th frame, we define two functions: $U_k$ and $V_k$:
\begin{eqnarray}
U_k & = & g(Y_{k-1},X) \\
V_k & = & h(Y_{k-1},\cdots,Y_1,X) 
\end{eqnarray}
where $U$ and $V$ model the short-term and long-term stimulus respectively for generating $Y_k$. These two functions are implemented as new layers in our architecture, which is discussed in subsequent sections. Therefore, we now model $P(Y|X)$ as $P(Y|U,V) = \prod^{n}_{i=1} P(Y_i|U_i,V_i)$.

\begin{figure*}[h]
\centering

\includegraphics[width=0.9\textwidth]{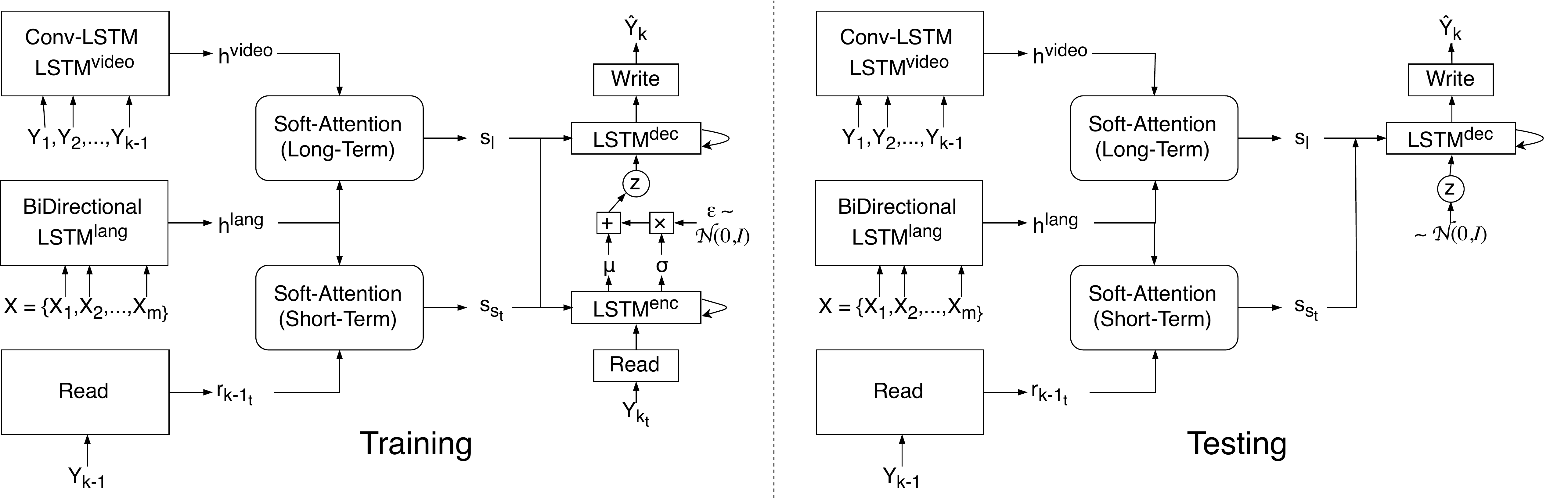}
\caption{Proposed network architecture for attentive semantic video generation with captions. $Y = \{Y_1,Y_2,\cdots,Y_k\}$ denotes the video frames generated by the architecture, while $X = \{X_1,X_2,\cdots,X_m\}$ denotes the set of words in the caption.}
\label{architecture}
\end{figure*}
$P(Y|U,V)$ is a complex multi-modal distribution that models the various possibilities of generating a video given a caption. For example, the same caption ``Man walking on grass'' can be generated with a man differing in height, face and other physiological attributes. Even the way the man walks can exhibit a wide range of variations. In order to capture this complex distribution $P(Y|U,V)$ and avoid an `averaging effect' over the various possibilities, we use a variational auto-encoder~(VAE)~\cite{kingma2013auto}. The VAE introduces a latent variable $z$ such that the likelihood of any given possibility is governed by the likelihood of sampling a particular value of $z$. So, we can define $P(Y|U,V)$ in terms of $z$ as (similar to ~\cite{kingma2013auto}), using normal distribution (denoted by $\mathcal{N}$):
\begin{equation}
P(Y|U,V) = \mathcal{N}(f(z,U,V),\sigma^2 * I)
\end{equation}

 
Using variational inference, a distribution $Q(z|Y,U,V)$ is introduced that takes values for $Y$, $U$ and $V$ in the training phase, and outputs an approximate posterior distribution over $z$ that is likely to generate the appropriate video given the long-term and short-term contexts derived from a given caption. Hence the variational lower bound is: 
\vspace{-6pt}
\begin{eqnarray}
\log~P(Y|U,V) =&KL(Q(z|U,V,Y) || P(z|U,V,Y))  \nonumber \\
&+\> E_{z \sim Q(z|U,V,Y)}[\log~P(Y,z|U,V) \nonumber \\
& -\>\log~Q(z|U,V,Y)] \nonumber \\
\geq & E_{z \sim Q(z|U,V,Y)} [\log~P(Y|U,V,z) \nonumber \\
& +\> \log~P(z|U,V) - \log~Q(z|U,V,Y)] \nonumber \\
 = & E_{z\sim Q(z|U,V,Y)}[\log~P(Y|U,V,z)) \nonumber \\
& -\> KL(Q(z|U,V,Y) || P(z|U,V)]
\end{eqnarray}
\noindent where $Q(z|U,V,Y)$ approximates the intractable posterior $P(z|U,V)$. $z$ is conditionally independent of $U$ and $V$ with $P(z|U,V) = P(z)$ and $P(z)$ is assumed to be $\mathcal{N}(0,I)$. 

One way to model $P(Y|U,V)$ for video generation is to introduce separate latent variables for each of the frames to be generated. Such an approach does not allow the model to scale and generate videos with arbitrary number of frames (since the number of latent variables will then be arbitrary). Therefore, we propose our model with just a single latent variable, $z$ to model $P(Y|U,V)$; the same latent variable is sampled recurrently to generate every subsequent frame. 
Figure~\ref{state-space} shows an illustration of the proposed model. During testing, we sample $z \sim \mathcal{N}\left(0,I\right)$ to get a sample (one video frame) from the distribution $P\left(Y|U,V\right)$. This is then continued recurrently, allowing us to generate videos with \textit{any user-defined number of frames}. 


 
\vspace{-2pt} 
\subsection{Latent Representation and Soft Attention for Captions}
\vspace{-6pt}
Similar to \cite{mansimov2015generating}, we first pass the given caption through a bi-directional $LSTM^{lang}$, as shown in Figure~\ref{architecture}, to generate a word-level latent representation for the caption, $h^{lang} = [h_1^{lang},h_2^{lang},\cdots,h_m^{lang}]$ with $h_i^{lang}$ denoting the latent representation for the $i^{th}$ word in the caption ($m$ being the total number of words in the caption).
Since $U$ and $V$ model different kinds of context for a video, it is reasonable to say that a caption can trigger $U$ and $V$ differently. For instance, in the example caption discussed so far, words like `grass' and `walking' that correspond to elements like background and motion in a video can trigger its long-term characteristics; while `man' whose posture changes in every frame can be responsible for the video's short-term characteristics. Therefore, we propose the network architecture to have a separate soft attention mechanism for each of the contexts over $h^{lang}$, the latent representation of the caption. This is described in the next few sections.

\vspace{-2pt}
\subsection{Modeling Long-Term Context using Attention}
\label{longattention}
\vspace{-6pt}
To model the long-term context, $V_k$, to generate the video frame $Y_k$, our network architecture consists of a convolutional LSTM, which we call $LSTM^{video}$ as shown in Figure~\ref{architecture}. We feed the frames $Y_{k-1},\cdots,Y_1$, generated so far by the network to $LSTM^{video}$, and compute $h^{video}$ which is the final latent representation learned by $LSTM^{video}$ after processing all the previous frames. This $h^{video}$ is then combined with $h^{lang}$ from the caption via a soft-attention layer to create the long-term context representation $s_l$ as shown in Figure~\ref{architecture}. The soft-attention mechanism receives $h^{lang}$ and $h^{video}$ as input, and works by learning a set of probabilities, $A = \{\alpha_1,\alpha_2,\cdots,\alpha_m\}$ corresponding to each word in the caption. The output of the soft-attention layer is then given by:
\vspace{-6pt}
\begin{eqnarray}
s_l &=& attention(h^{lang},h^{video}) \nonumber \\
& =& \alpha_1 h_1^{lang} + \alpha_2 h_2^{lang} + \cdots + \alpha_m h^m_{lang}
\label{eq10and11}
\end{eqnarray}
where:
\vspace{-6pt}
\begin{equation}
\alpha_i = \frac{\exp\left(v^T \tanh(u h_i^{lang} + w h^{video} + b)\right)}{\sum_{j=1}^{m} \exp\left(v^T \tanh(u h_j^{lang} + w h^{video} + b)\right) \nonumber }
\label{eq12}
\end{equation}
where $v$, $u$, $w$ and $b$ are the network parameters.

\subsection{Modeling Short-Term Context using Attention and Frame Generation}
\vspace{-6pt}
When a frame is generated in a single pass, even though the model might be able to preserve the overall motion in the frame, the various objects in the scene suffer from blurriness and distortion. In order to overcome this drawback of one-shot frame generation, we model $U_k$ and propose our frame generator to have a differentiable recurrent attention mechanism resembling \cite{graves2014neural} and \cite{gregor2015draw} to generate frame $Y_k$ in $T$ timesteps (as shown in Figure~\ref{architecture}). The attention mechanism comprises of a grid of Gaussian filters, whose grid center, variances and stride are learnt by the network (as in \cite{gregor2015draw}). For every timestep $t$, we read a glimpse, $r_{k-1}$, from the previous frame, $Y_{k-1}$, pass it through a small convolutional network (2 convolutional layers with one fully connected layer) and combine it with $h^{lang}$ via soft-attention (similar to Section~\ref{longattention}) to create the short-term context representation $s_{s_t}$. Simultaneously, we also read a glimpse, $r_t$, from $Y_{k_{t-1}}$ (also passed through a similar convolutional network), denoting the frame $Y_k$ generated after $t-1$ timesteps.


$r_t$ and $s_{s_t}$ along with $s_l$ Section~\ref{longattention} are encoded by $LSTM^{enc}$ to learn the approximate posterior $Q$. $z$ is then sampled from $Q$ and decoded using $LSTM^{dec}$, whose output is sent to a similar deconvolutional network before passing to a write function to generate content in a region of interest (learned by the attention mechanism) on the current frame of interest, $Y_k$. This is different from \cite{gregor2015draw}, and we found this to be important in generating videos of better quality. The use of such an LSTM autoencoder architecture in our model ensures that the region to be attended to next is conditioned on the regions that have been attended so far. The information written on different regions in the frame is accumulated over $T$ timesteps to generate a single frame $Y_k$. The following equations explain these steps: 
\vspace{-6pt}
\begin{eqnarray}
r_{{k-1}_t} &=& conv(read(Y_{k-1})) \\
r_t &=& conv(read(Y_{k_{t-1}})) \\
s_{s_t} &=& attention(h_{lang}, [r_{{k-1}_t}, h^{dec}_{t-1})]\\
\label{h_enc}
h^{enc} &=& LSTM^{enc}(r, s_s, s_l, h^{dec}_{t-1})\\
z & \sim & Q(z|h^{enc})\\
\label{h_dec}
h^{dec}_{t} &=& LSTM^{dec}(z, s_s, s_l, h^{dec}_{t-1})\\
Y_{k_{t}} &=& Y_{k_{t-1}} + write(deconv(h^{dec}_{t}))
\end{eqnarray}
A major advantage of such a recurrent attention mechanism, in the context of this work, is that the network learns a single distribution that can distinguish between different elements of a frame, and attend to them in each time step. This further enables the network to dynamically combine these elements during inference to effectively generate frames.

\subsection{Loss Function}
\vspace{-6pt}
As mentioned in the previous sections, we take $P(z|U,V) = P(z) \sim \mathcal{N}(0,I)$. This simplifies the empirical variational  bound to (note that we minimize the negative of the bound, as in VAEs~\cite{kingma2013auto}):
\vspace{-6pt}
\begin{equation}
\small
\mathcal{L} = - \Bigg(\frac{1}{S}\sum^{S}_{s=1}{log~P(Y|U,V,z^s)} - KL(Q(z|U,V,Y) || P(z|U,V))\Bigg) 
\end{equation}
with 
\vspace{-6pt}
\begin{equation}
\small
KL(Q(z|U,V,Y) || P(z|U,V)) = 
\frac{1}{2} \Bigg( \sum\limits_{t=1}^T \mu_t^2 + \sigma_t^2 - \log \sigma_t^2 \Bigg) - T/2
\end{equation}
where $t$ denotes the time-step over which the frames are generated as before, and $z^s$ denotes the $s^{th}$ sample taken from the $z$ distribution among a set of total $S$ samples which are used to compute the likelihood term. The negative likelihood term, which is also the reconstruction loss, is computed as the binary pixel-wise cross-entropy loss between the original video frame and the generated frame. All the losses here are calculated frame-wise.

\vspace{-6pt}
\section{Experiments and Results}
\vspace{-6pt}
We evaluated the proposed model on datasets of increasing complexity\footnote{All the codes, videos and other resources are available at \url{https://github.com/Singularity42/cap2vid}}. We first created a variant of Moving MNIST dataset (similar to \cite{srivastava2015unsupervised,kalchbrenner2016video}) with videos depicting a single digit moving across the frame. Each video has a set of 10 frames, each of size $64 \times 64$. We added the $28 \times 28$ sized images of digit from the original MNIST dataset to each of the frames and varied the initial positions to make the digit move either up-and-down or left-and-right across the frames. We then captioned each video based on the digit and its motion. For instance, a video with caption `digit 2 is going up and down' contains a sample of 2 from MNIST moving up and down in the video. We similarly created a Two-Digit Moving MNIST dataset similar to~\cite{srivastava2015unsupervised,kalchbrenner2016video} where each video contains two randomly chosen digits moving across the frames going either left-and-right or up-and-down independently, giving us a dataset containing 400 combinations. Examples of the captions in this dataset are `digit 1 is moving up and down and digit 3 is moving left and right'.

To evaluate the proposed network's performance on a more realistic dataset, we used the KTH Human Action Database~\cite{schuldt2004recognizing} which consists of over 2000 video sequences comprised of 25 persons performing 6 actions (walking, running, jogging, hand-clapping, hand-waving and boxing). We used the video sequences of walking, running and jogging for our evaluation because in each of these actions, a person was going either right-to-left or left-to-right which allowed us to introduce a sense of `direction' in the video context, and study the proposed model. We uniformly sampled from the given video sequence and generated our dataset with each video having 10 frames of size $120 \times 120$. For each of these videos, we manually created captions such as `person 1 is walking left-to-right' or 'person 3 is running right-to-left'. The information on person number is obtained from the metadata accompanying the dataset. The advantage of using the KTH dataset in our context is that the same set of people perform all the actions as opposed to other action datasets (such as UCF-101) where the people performing the actions changes. This allows us to add appropriate captions to the dataset, and study and validate the performance of our model.

We further performed experiments for unsupervised video generation using our model without captions. As it is related to earlier efforts, we show results on UCF-101 dataset~\cite{soomro2012ucf101} to be able to compare with earlier efforts. We uniformly sampled the videos from UCF-101 dataset to generate video sequences for our training dataset with each video having 10 frames each of size $120\times120\times3$. Important thing to note here is that although we are training our network over videos of 10 frames, we can generate videos with any number of frames (shown in Section \ref{subsec_unsupervised_results}).

We note that in order to generate the first frame of the video for a caption, we prefixed videos from all datasets with a start-of-video frame. This frame marks the beginning of every given video. It contains all $0s$ resembling the start-of-sentence tag used to identify the beginning of a sentence in Natural Language Processing~\cite{sutskever2014sequence}.

\begin{figure*}[h]
\centering
\includegraphics[width=0.7\textwidth]{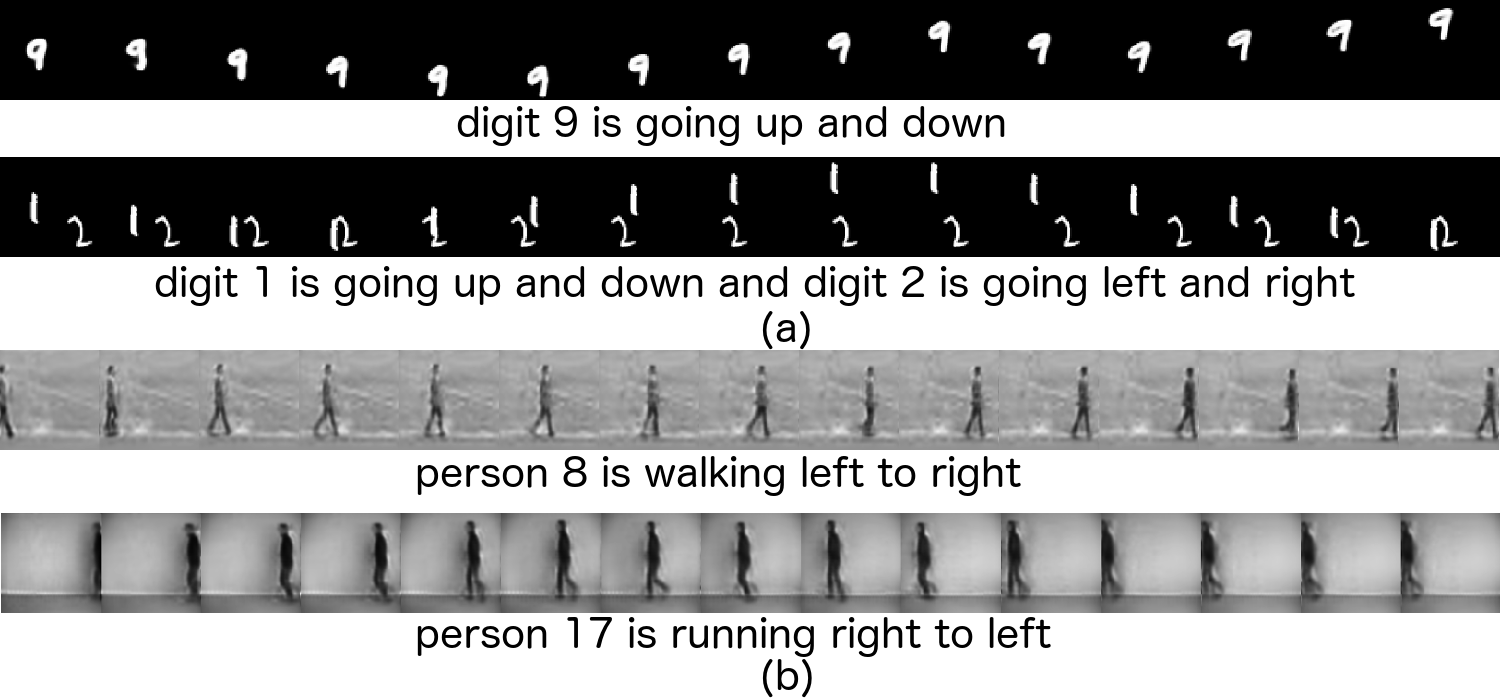}
\caption{The results of our network for different datasets to generate videos for a single caption. Number of frames generated is 15.}
\label{mnist-kth-basic}
\end{figure*}


\subsection{Results on Generation with Captions}
\begin{figure}[h]
\centering

\includegraphics[width=0.9\columnwidth]{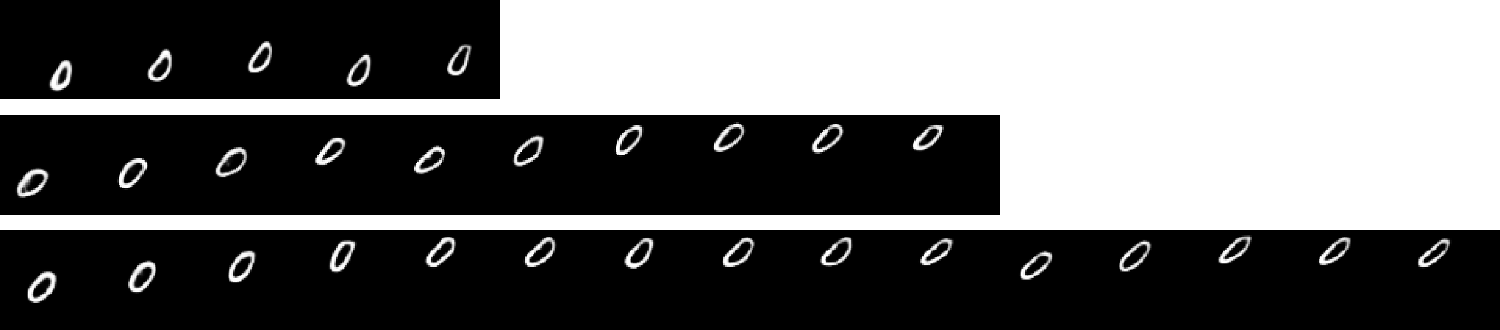}
\caption{Variable length video generation for caption `digit 0 is going up and down'.}
\label{variable-length}
\end{figure}

\vspace{-6pt}
The results on the Moving MNIST and KTH datasets are shown in Figure~\ref{mnist-kth-basic}, and illustrate the rather smooth generations of the model. (The results of how one frame of the video is generated over $T$ time-steps is included in the supplementary material due to space constraints.) In order to test that the network is not memorizing, we split the captions into a training and test set. Therefore, if the training set contains the video having caption as `digit 5 is moving up and down', the model at test-time is provided with `digit 5 is moving left and right'. Similarly, in the KTH dataset, if the video pertaining to the caption `person 1 is walking from left-to-right' belongs to the training set, the caption `person 1 is walking right-to-left' is included at test time. We thus ensure that the network makes a video sequence from a caption that it has never seen before. The results show that the network is indeed able to generate good quality videos for unseen captions. We infer that it is able to dissociate the motion information from the spatial information. In natural datasets where the background is also prominent, the network selectively attends to the object over the background information. The network achieves this without the need of externally separating the object information from the background information~ \cite{vondrick2016generating} or external motion trajectories~\cite{walker2016uncertain}. Another observation is that the object consistency is maintained and the long-term fluidity of the motion is preserved. This shows that the long-term and the short-term context is effectively captured by the network architecture and they both work in `coordination' to generate the frames. Also, as mentioned earlier, methodology ensures that a video can be generated with any number of frames. An example of a generation with variable length of frames is shown in Figure~\ref{variable-length}.

\subsection{Results with Spatio-Temporal Style Transfer}
\vspace{-6pt}
\begin{figure*}[h]
\centering

\includegraphics[width=0.7\textwidth]{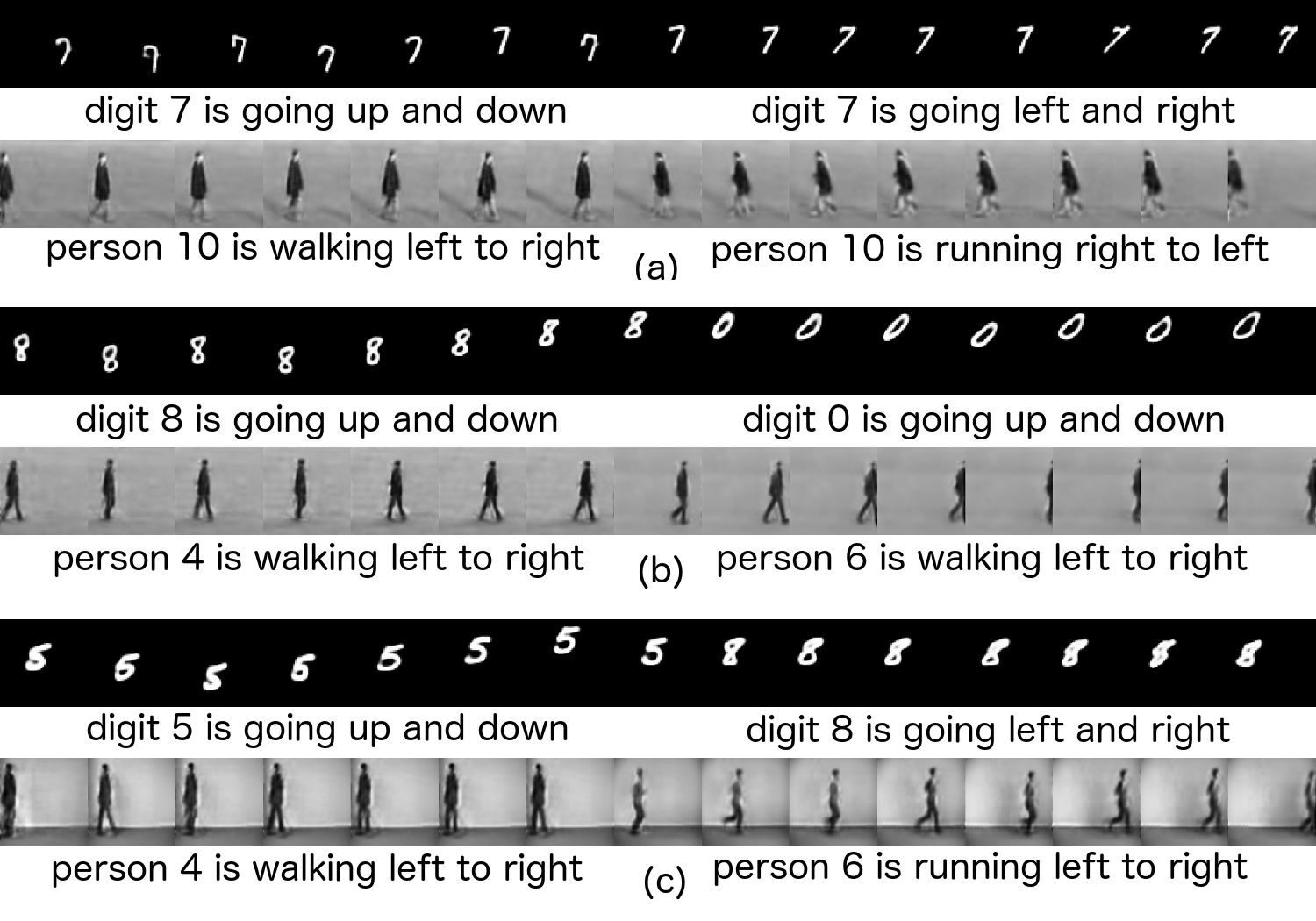}
\caption{Spatio-Temporal style transfer. First caption generates for $7$ frames. Second caption continues the generation from the $8^{th}$ frame.}
\label{spatio-temporal-transfer-results}
\end{figure*}
We further evaluated the capability of the network architecture to generate videos where the captions are changed in the middle. Here, we propose two different settings: (1) \textit{Action Transfer}, where the information of the motion that the object (i.e, digit/person) is performing is changed midway during the generation; and (2) \textit{Object Transfer}, where the object information (i.e, digit/person) is changed midway during the generation. During action transfer, we expect that the object information remains intact but the motion that is being performed by the object changes; and during object transfer, we expect a change in the object information with respect to what has been generated so far. Results for action and object transfer can be seen in Figures~\ref{spatio-temporal-transfer-results}(a) and~\ref{spatio-temporal-transfer-results}(b) respectively. We also go a step ahead and perform both \textit{Action Transfer} and \textit{Object Transfer} together as shown in Figure~\ref{spatio-temporal-transfer-results}(c). To test the robustness of the network, we ensured that the second caption used in this setup was not used for training the network. 

We note here that when the spatio-temporal transfer happens, the object position remains the same in all the results, and the object from the second caption continues its action from exactly the same position. This is different from the case when the video is freshly generated using a caption since then the object can begin its motion from any arbitrary position. Moreover, the network maintains the context of the video while changing the object or action. For example, in Figure~\ref{spatio-temporal-transfer-results}(c), the digit 5 with a certain stroke width and orientation changes to a digit 8 with the same stroke width and orientation. Similarly, in the natural dataset the type of the background and its illumination remains the same. The preservation of motion and context as well as the position is a crucial result in showcasing the ability of the network to maintain the long-term and short-term context in generating videos even when the caption is changed in the middle.

\begin{figure}[h]
\centering
\includegraphics[width=0.9\columnwidth]{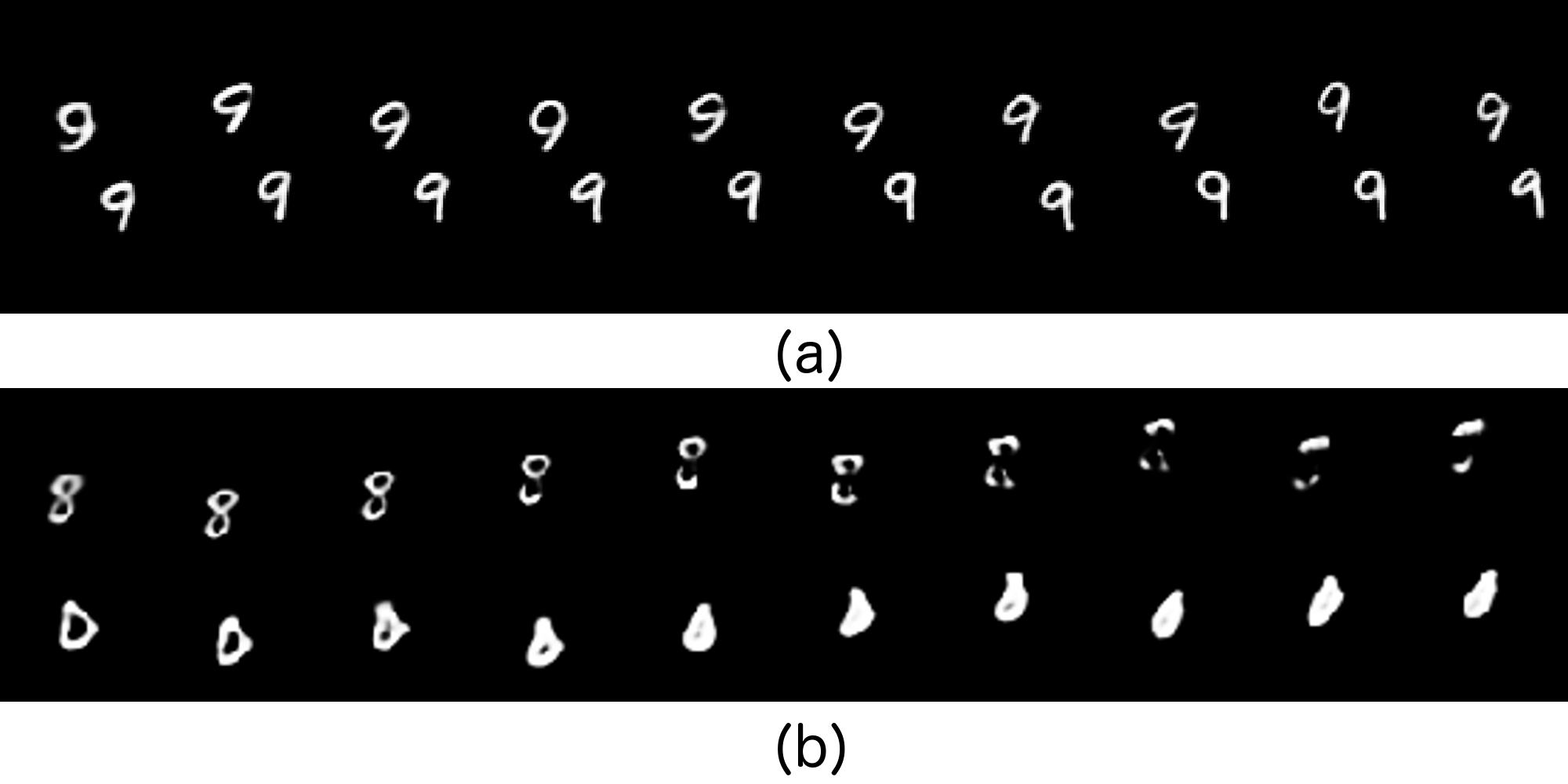}
\caption{(a) shows videos when generated without captioning on Long-term context. (b) shows videos when generated without captioning on Short-term context.}
\label{ablation}
\end{figure}

\begin{table}
\small
\begin{tabular}{|l|c|c|c|}
\hline
\hline
\multicolumn{1}{|c|}{\textbf{Captioning Experiment}}                                        & \textbf{LL}    & \textbf{KL}    & \textbf{Total Loss} \\ \hline
\begin{tabular}[c]{@{}l@{}}Without Captioning on \\ Long-Term Context\end{tabular}  & 65.76 & 17.97 & 83.73      \\ 
\hline
\hline
\begin{tabular}[c]{@{}l@{}}Without Captioning on \\ Short-term Context\end{tabular} & 72.53 & 13.56 & 86.09      \\ \hline
\begin{tabular}[c]{@{}l@{}}With Captioning on Long-\\ and Short-Term Context\end{tabular}         & \textbf{63.55} & \textbf{11.84} & 
\textbf{75.39}      \\ 
\hline
\hline
\end{tabular}
\caption{Quantitative comparison of loss at the end of training for different ablation experiments on captioning.}
\label{ablation-table}
\end{table}
\subsection{Ablation Studies}
\vspace{-6pt}
The key contribution of this work has been to condition the generation of the videos using long-term context ($s_l$) and short-term context ($s_s$). We perform an ablation study over the importance of each of these contexts, by removing them one at a time and training the network. We ensure that all the other parameters between the two networks remain the same. The results of removing long-term context are shown in Figure~\ref{ablation}~(a), where the caption is `digit 9 moving up and down'. Compared to Figure~\ref{mnist-kth-basic}(a), there are two distinct effects that can be observed: (1) either the object characteristics (such as the shape of the digit 9) changes over time; or (2) the object starts to oscillate (instead of following the specified motion) because it has failed to capture the long-term fluid motion over the video, thus supporting the need for $s_l$ in maintaining coherence between the frames. 

The results of removing short-term context are shown in Figure~\ref{ablation}~(b). Here we notice a significant deterioration in the object quality. We infer that as the frame generation takes place over a  number of time-steps, it is essential that the model, while generating a frame in a given time-step, receives a strong local temporal stimulus from glimpses of the previous frame. Table~\ref{ablation-table} further shows a quantitative analysis of these experiments, which corroborate that the proposed approach of using a long-term and short-term context together helps learn a better model for video generation.

\begin{figure}[h]
\centering
\includegraphics[width=0.9\columnwidth]{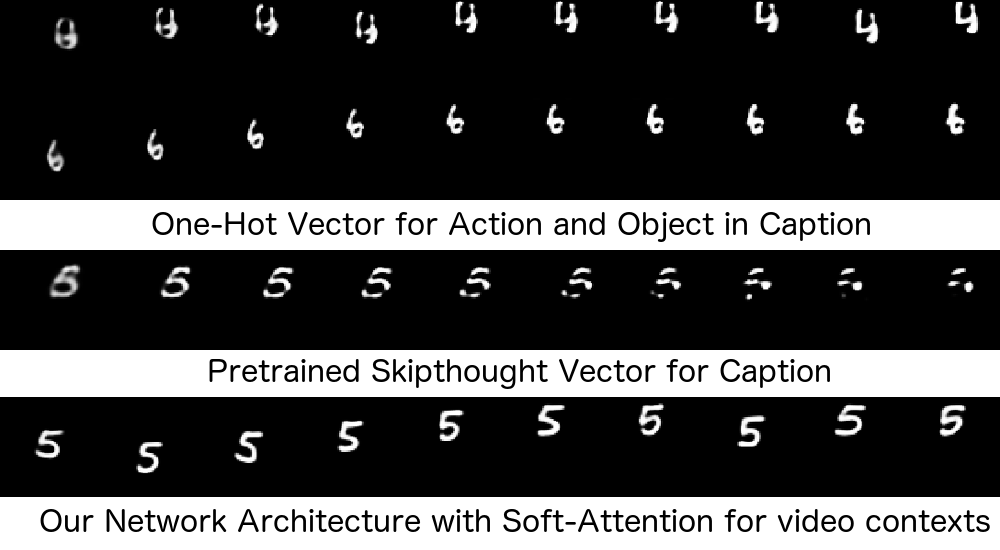}
\caption{Videos generated using different approaches of conditioning over test-set caption `digit 5 is going up and down'}
\label{caption_ablation}
\end{figure}
In order to further study the usefulness of the attention mechanism we have on the caption representation, we compared our approach with other ways to condition on the caption information. We first trained our network where we conditioned the frame generation directly on one-hot vectors created separately from the action and object information present in the captions. We also performed an experiment where we conditioned on latent vectors obtained by passing the caption through a pre-trained model of skipthought vectors~\cite{kiros2015skip}. The results of the comparison are shown in Figure~\ref{caption_ablation}. It can be observed that the one-hot vector approach did not respond to the caption at all and generated random video samples. The approach with pre-trained skipthought vectors did match the object information but couldn't associate it with the correct motion. In comparison, our methodology is able to perform extremely well for the unseen caption.



\begin{figure}[h]
\centering

\includegraphics[width=0.9\columnwidth]{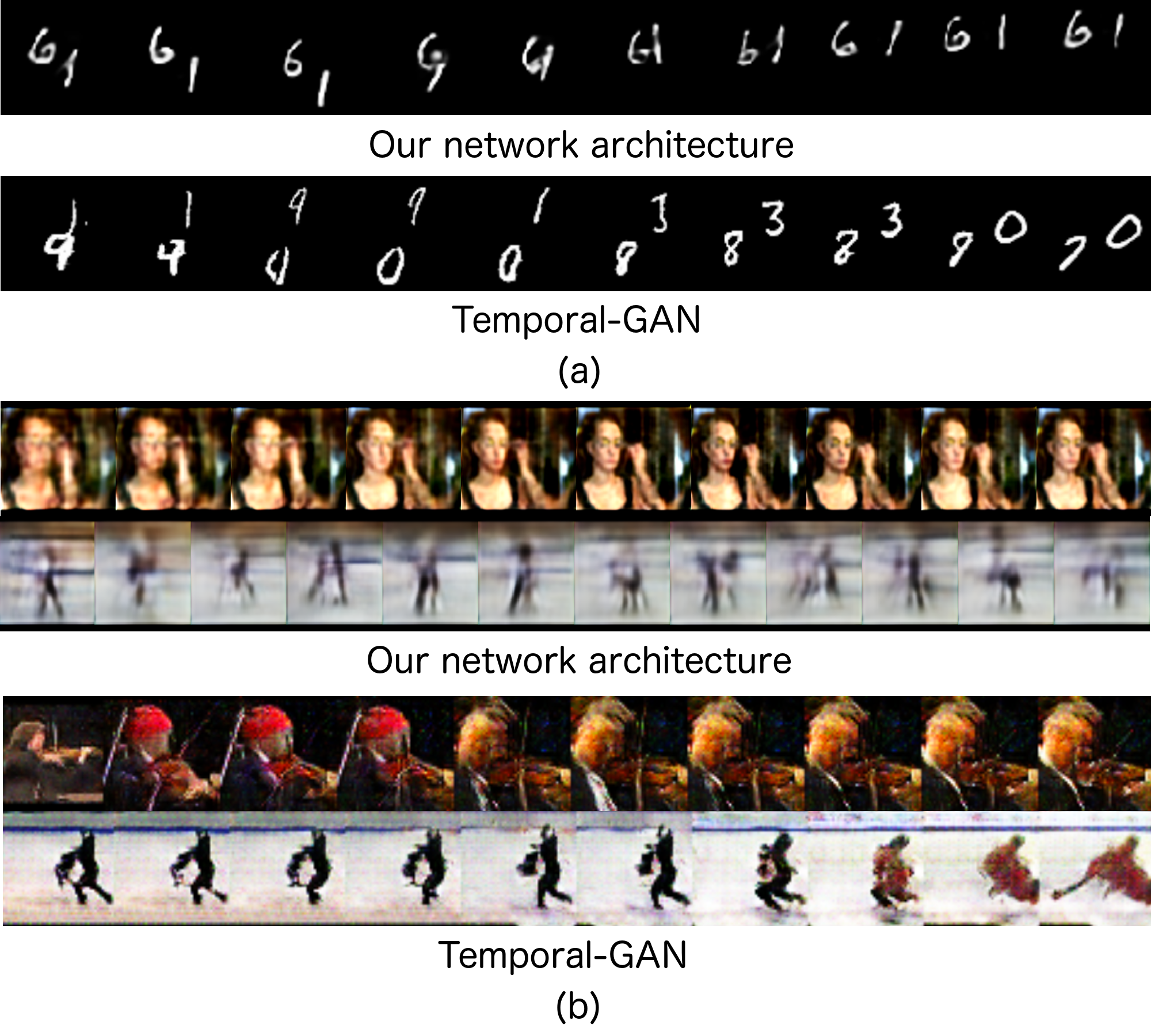}
\caption{Results of unsupervised video generation and comparison with results from Temporal-GAN~\cite{saito2016temporal}.}
\label{unsup}
\end{figure}

\vspace{-8pt}
\subsection{Unsupervised Video Generation}
\label{subsec_unsupervised_results}
\vspace{-6pt}

The proposed architecture also allows videos to be generated in an unsupervised manner without captions by making some modifications. In order to do this,we remove $s_s$ and $s_l$ and the frame generation is conditioned directly on $r_{{k-1}_t}$ and $h^{video}$ changing equations~\ref{h_enc} and \ref{h_dec} to:
\begin{eqnarray}
h^{enc}_{t} &=& LSTM^{enc}(r, r_{{k-1}_t}, h_{video}, h^{dec}_{t-1})\\
h^{dec}_{t} &=& LSTM^{dec}(z, r_{{k-1}_t}, h_{video}, h^{dec}_{t-1})
\end{eqnarray}
We show the effectiveness of our architecture to generate unsupervised videos by training our network on Two-Digit Moving MNIST dataset and UCF-101 datasets. As mentioned earlier, we used UCF-101 in this experiment to be able to compare with earlier work. (Also, UCF-101 was not used in other experiments, since it does not have captions and it's not trivial to add captions for this dataset.) Results can be seen in Figure~\ref{unsup} along with a few generations presented by \cite{saito2016temporal} in their paper for comparison. We can infer that even in the unsupervised setting, our method preserves the object information throughout the video, whereas in \cite{saito2016temporal}, all the generated videos seem to lose the objectness towards the later part of the video. 

\vspace{-6pt}
\subsection{Quantitative Results}
\vspace{-6pt}
\begin{table}[]
\small
\centering
\begin{tabular}{|l|c|}
\hline
\hline
\multicolumn{1}{|c|}{Experiment}         & Accuracy (\%) \\ 
\hline
\hline
Training: Generated, Testing: Original & 62.36         \\ \hline
Training: Original, Testing: Generated & 68.24        \\ \hline
Training: Both, Testing: Both          & \textbf{70.95}         \\ 
\hline
\hline
\end{tabular}
\caption{Accuracy on action classification performed with feature vectors from original and generated videos on the KTH dataset.}
\label{action-table}
\end{table}
\noindent \textbf{Data programming:} Since our network can generate caption based videos, it can even be employed to artificially create a labeled dataset~\cite{ratner2016data}. So, we evaluated the real and generated videos of KTH dataset for action recognition. We first trained an SVM on features extracted from the generated videos with action labels corresponding to the captions, and tested the model on the real videos. We then performed the experiment vice versa. To ensure an independent unbiased analysis, we extracted feature vectors using a 3D spatio-temporal convolutional network~\cite{tran2015learning} trained on Sports 1M Dataset ~\cite{KarpathyCVPR14}. As shown in Table~\ref{action-table}, we can observe that the accuracy for the two settings is comparable. In fact, the highest is achieved on mixing the two datasets. We infer that the generated videos are able to fill up the gaps in the manifold and thus oversampling the original set of video with them can help improve the training on a supervised learning task. (We note that these models were not finetuned to achieve the best possible accuracy on the dataset, but only used for comparison against each other.)

\noindent \textbf{User Study: }In order to assess how people perceive our generated results, we performed a user study. 10 generated and 10 real videos having the same captions for each dataset were shown to 24 subjects who were asked to rate the videos as generated or real. The results are shown in Table~\ref{PP analysis}. It can be clearly observed that the percentage of generated videos considered real is very close to that of real videos. This suggests that our network is able to generate new videos highly similar to the real ones.

\begin{table}[ht]
 \small
 \centering
 \setlength\tabcolsep{3pt}
 \newcommand{\specialcell}[2][c]{%
  \begin{tabular}[#1]{@{}c@{}}#2\end{tabular}}
 \begin{tabular}{|c|c|c|c|c|c|c|}
 \hline
 \hline
Datasets & \multicolumn{2}{|c|}{\specialcell{One Digit MNIST}} & \multicolumn{2}{|c|}{\specialcell{Two Digit MNIST}} & \multicolumn{2}{|c|}{\specialcell{KTH Dataset}}\\
 \hline
Video Type &Gen.&Real&Gen.&Real&Gen.&Real\\
 \hline
 \hline
 \specialcell{\% videos\\considered real}&83.33\%&93.75\%&78.53\%&89.15\%&75.94\%&92.45\%\\
 \hline
 \hline
 \end{tabular}
 \caption{\textbf{User Study} Results showing the percentage of videos considered real by people.}
 \label{PP analysis}
 \end{table}
\vspace{-8pt}
(We also performed studies on the task of future frame prediction, but are unable to include it due to space constraints. These are included in the supplemental material.)

\vspace{-2pt}
\section{Conclusion}
\vspace{-6pt}
In summary, we proposed a network architecture that enables variable length semantic video generation using captions, which is the first of its kind. Through various experiments, we conclude that our approach, which combines the use of short-term and long-term spatiotemporal context, is able to effectively to generate videos on unseen captions maintaining a strong consistency between consective frames. Moreover, the network architecture is robust in transferring spatio-temporal style across frames when generating multiple frames. By learning using visual context, the network is even able to learn a robust latest representation over parts of videos, which is useful for tasks such as video prediction and action recognition. We did observe in our experiments that our network does exhibit some averaging effect when filling up the background occasionally. We infer that since the recurrent attention mechanism focuses primarily on objects, the approach may need to be extended with a mechanism to handle the background via a separate pathway. Future efforts in this direction will assist in bridging this gap as well as extending this work to address the broader challenge of learning with limited supervised data.

{\small
\bibliographystyle{ieee}
\bibliography{egbib}
}

\newpage

\section{Supplementary Material}

\begin{figure*}[p]
\centering
\includegraphics[width=\textwidth]{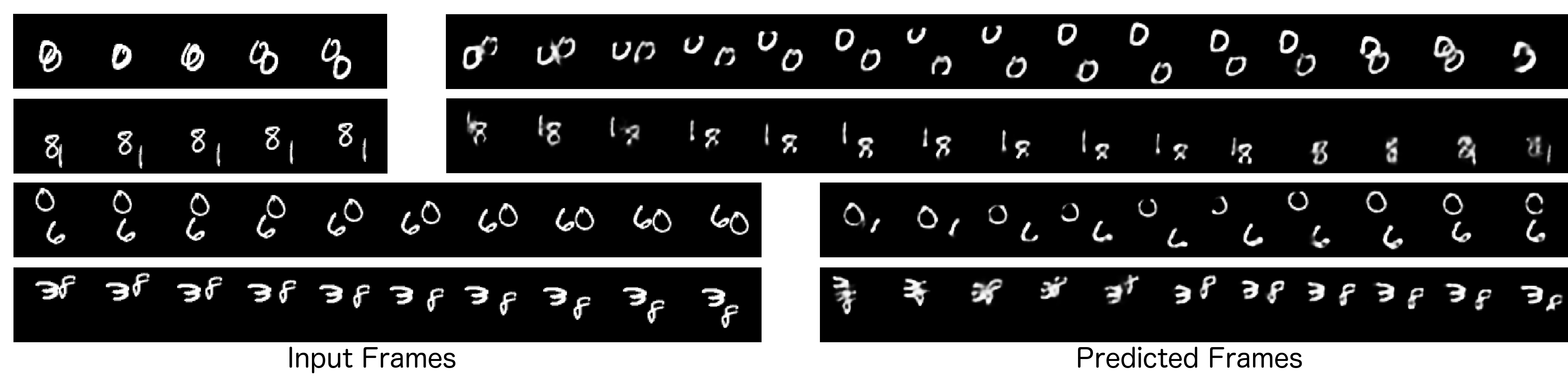}
\caption{Next frame prediction on Moving MNIST with variable number of frames specified for input and prediction.}
\label{future-frame-prediction}
\end{figure*}

We herewith present additional results using the proposed methodology, which could not be included in the main paper due to space constraints.

\paragraph{Future frame prediction} We evaluated our approach on the task of next frame prediction. We took the Two-Digit Moving MNIST dataset that we used for other experiments and fed the first few frames to the model, asking it to predict the next set of frames. We trained our network in two settings: (1) When the number of input video frames is $10$ and the network is allowed to predict the subsequent $10$ frames; and (2) When the number of input video frames is 
$5$ and the network is allowed to predict the subsequent $15$ frames (in order to showcase the network's ability to generate variable lengths of frames). The results are shown in Figure~\ref{future-frame-prediction}. It can be observed that our model performs well on this task. Although there are minor distortions observed when the two digits overlap, the network is robust enough not to perpetuate these distortions and it maintains the overall spatio-temporal consistency across the frames generated. We note that earlier efforts on future frame prediction (such as \cite{kalchbrenner2016video,srivastava2015unsupervised}) only predict a pre-defined fixed number of frames.

\begin{figure*}[h]
\centering
\includegraphics[trim={0 4cm 0 0}, clip=true, width=\textwidth]{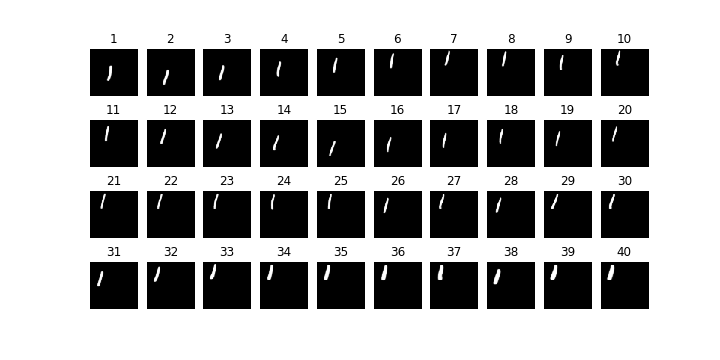}
\caption{Video generated for Moving MNIST for caption 'digit 1 is going up and down' for $30$ frames, showing the ability of the network to generate a video with an arbitrary number of frames.}
\label{long_generation_mnist}
\end{figure*}

\paragraph{Variable length video generation} Figure~\ref{long_generation_mnist} demonstrates the ability of the network to generate videos containing a relatively large number of frames by generating a video containing $30$ frames. 

\begin{figure*}[h]
\centering

\includegraphics[width=0.8\textwidth]{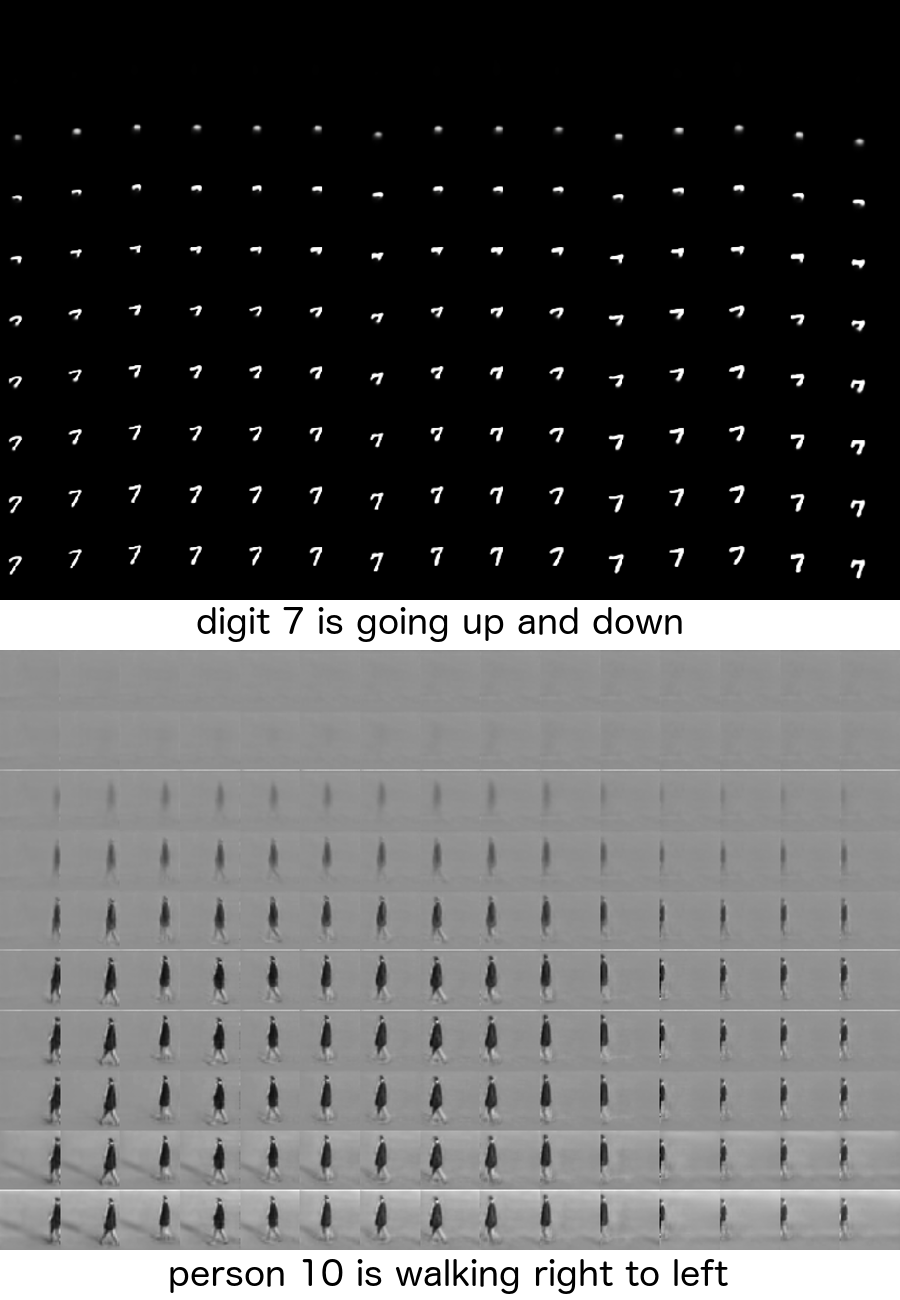}
\caption{Video generation for Moving MNIST and KTH dataset. Here $T=10$ and $n=15$. The frames are generated one at a time conditioned on all the previously generated frames.}
\label{video_formation}
\end{figure*}

\begin{figure*}[h]
\centering
\includegraphics[width=\textwidth]{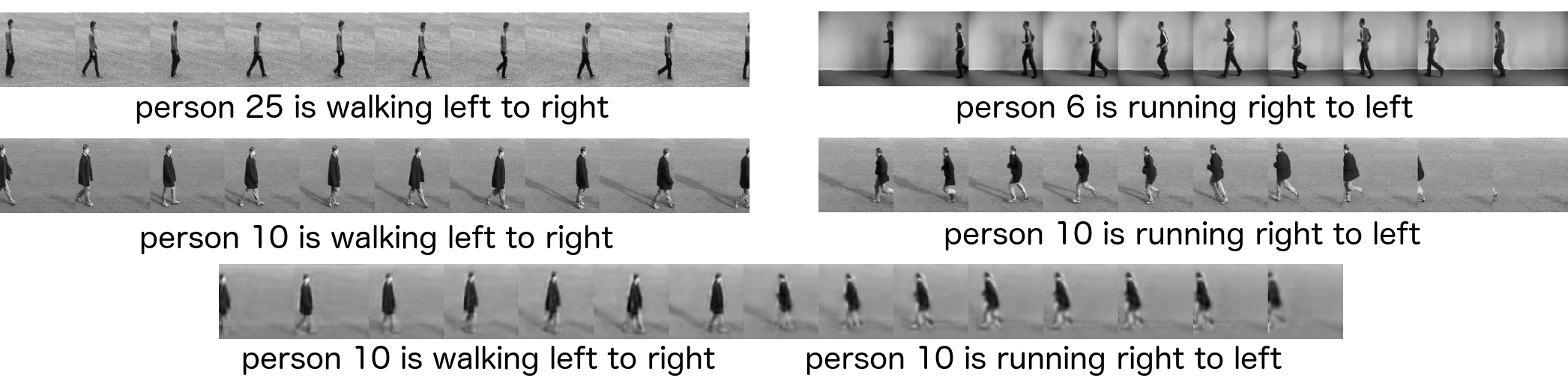}
\caption{To demonstrate that the network understands the object information, the top row shows the generated video belonging to persons 25 and 6. The second row shows videos of person 10. In the third row, we show the video generated for the captions `person 10 is walking left to right' and `person 10 is running right to left' (action transfer during the video).}
\label{kth_data_explanation}
\end{figure*}

\paragraph{Example of video formation over $T$ timesteps} Figure~\ref{video_formation} shows how the network generates the various frames of the video over $T$ timesteps. Here we have taken $T=10$ and the number of frames generated, $n=15$. The frames are generated one at a time conditioned on all the previously generated frames. Figure~\ref{video_formation} shows them together to make it easier to visualize the motion in the video. It is evident that the model gradually generates each video, by conditioning the content on each frame on previously generated content. 
An interesting observation in the generation on the KTH dataset is that towards the end, the video shows a person leaving the field of view of the camera. This happens because the training data for person 10 has such frames towards the end of certain videos too (as seen in Figure \ref{kth_data_explanation}). 

\paragraph{Analysis of results on the KTH dataset} In order to assert that our network architecture is indeed able to generate the correct person given the person index in case of the KTH dataset, we show in Figure~\ref{kth_data_explanation}, three rows of videos. The top row depicts videos from the dataset of different persons. The second row shows videos from person 10, and the third row shows the generation of our model in the action transfer setting. It is evident that the generated video, conditioned on the captions including person 10, indeed corresponds to person 10, and not of any other person.

\begin{figure*}[h]
\centering
\includegraphics[width=\textwidth]{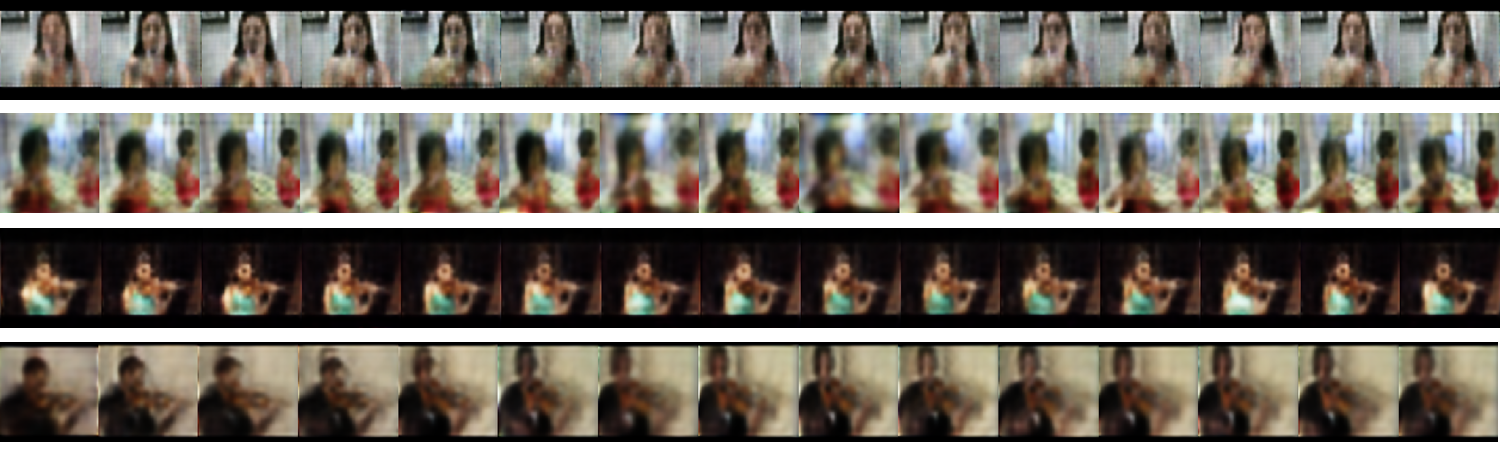}
\caption{Unsupervised video generation for the UCF dataset. The first two rows show videos where people are brushing their teeth, while the next two rows show people playing violin.}
\label{ucf_results}
\end{figure*}

\paragraph{More results on the UCF dataset} Figure~\ref{ucf_results} shows more results on the UCF dataset, which were generated in an unsupervised manner. The videos generated consist of $15$ frames, and show the effectiveness of the proposed framework despite the UCF dataset being very difficult to model.

\end{document}